\documentclass[conference,letter]{IEEEtran}  % Comment this line out if you need a4paper

\usepackage{amsmath,amsfonts}
\usepackage[ruled,vlined,linesnumbered]{algorithm2e}
\usepackage{array}
\usepackage[caption=false,font=normalsize,labelfont=sf,textfont=sf]{subfig}
\usepackage{textcomp}
\usepackage{stfloats} 
\usepackage{url}
\usepackage{verbatim}
\usepackage{graphicx}
\usepackage{cite}

\usepackage{amssymb}
\usepackage{orcidlink}
\usepackage{overpic}

\usepackage{graphicx} % for pdf, bitmapped graphics files
\usepackage{tabularx}
\usepackage{longtable}
\usepackage{booktabs}
\usepackage{lscape}
\usepackage{multirow}

\usepackage{tikz}

\usepackage{xcolor}

\usetikzlibrary{arrows.meta, fit, positioning, backgrounds, calc, shapes.geometric}

\definecolor{Cactor}{RGB}{30, 100, 185}
\definecolor{Ccritic}{RGB}{195, 95, 15}
\definecolor{Cenv}{RGB}{28, 135, 65}
\definecolor{Cbuf}{RGB}{105, 55, 175}
\definecolor{Calgo}{RGB}{55, 55, 65}
\definecolor{Cpd}{RGB}{175, 40, 85}
\definecolor{Cpanel}{RGB}{228, 238, 252}
\definecolor{Cpborder}{RGB}{155, 185, 225}
\definecolor{Csem}{RGB}{106, 27, 154}      % semantic layer (Fig. 3 roadmap update)

\newcommand{\best}[1]{\textbf{\textcolor[HTML]{B0006C}{#1}}}
\newcommand{\second}[1]{\underline{#1}}

\tikzset{
  blk/.style={rectangle, rounded corners=4pt, minimum width=#1,
    minimum height=0.72cm, draw, very thick, align=center},
  blk/.default=2.4cm,
  actor/.style  ={blk=2.4cm, fill=Cactor!13, draw=Cactor,
                  font=\small\bfseries, text=Cactor},
  critic/.style ={blk=2.7cm, fill=Ccritic!13, draw=Ccritic,
                  font=\small\bfseries, text=Ccritic!90!black},
  envblk/.style ={blk=2.5cm, fill=Cenv!13,    draw=Cenv,
                  font=\small\bfseries, text=Cenv!90!black},
  bufblk/.style ={blk=2.7cm, fill=Cbuf!13,    draw=Cbuf,
                  font=\small\bfseries, text=Cbuf!90!black},
  algoblk/.style={blk=2.4cm, fill=Calgo!10,   draw=Calgo,
                  font=\small\bfseries, text=Calgo},
  iobox/.style  ={blk=2.1cm, fill=white, draw=gray!55, rounded corners=3pt,
                  font=\scriptsize\bfseries, text=gray!65!black,
                  minimum height=0.58cm},
  pdblk/.style  ={blk=1.7cm, fill=Cpd!10, draw=Cpd!75,
                  font=\scriptsize\bfseries, text=Cpd!80!black,
                  minimum height=0.58cm},
  intblk/.style ={blk=1.45cm, fill=Cpd!7, draw=Cpd!55,
                  font=\scriptsize\bfseries, text=Cpd!75!black,
                  minimum height=0.58cm},
  lbl/.style    ={font=\scriptsize, text=gray!65!black},
  A/.style ={-{Stealth[length=5pt,width=3.2pt]}, thick},
  Ab/.style={A, Cactor!80},
  Ag/.style={A, Cenv!80},
  Ao/.style={A, Ccritic!90!black},
  Ap/.style={A, Cbuf!80},
  Ad/.style={A, dashed, gray!55},
}

\begin{document}

\title{Semantic-Aware Autonomous Exploration for UAVs in Unknown Indoor Environments}

\author{
\IEEEauthorblockN{Duc-Thien Nguyen\textsuperscript{1}, Ngoc Minh Do\textsuperscript{1}, $^*$Xiem HoangVan\textsuperscript{1}\orcidlink{0000-0002-7524-6529}, Thanh Nguyen Canh\textsuperscript{1, 2}~\orcidlink{/0000-0001-6332-1002}, }
\IEEEauthorblockA{\textsuperscript{1}University of Engineering and Technology, Vietnam National University, 10000, Hanoi, Vietnam.}
\IEEEauthorblockA{\textsuperscript{2}School of Information Science, Japan Advanced Institute of Science and Technology, Nomi, 923-1211, Ishikawa, Japan.}
% \IEEEauthorblockA{\textsuperscript{3}Department of Robotics, Hanyang University, Ansan, 15588, Gyeonggi, Korea.}
\IEEEauthorblockA{*Corresponding author: \tt xiemhoang@vnu.edu.vn}
}

% The paper headers (template text removed; not required for conference papers)

\maketitle

\makeatletter
\setlength{\@fptop}{0pt}
\makeatother

%%%%%%%%%%%%%%%%%%%%%%%%%%%%%%%%%%%%%%%%%%%%%%%%%%%%%%%%%%%%%%%%%%%%%%%%%%%%%%%%
\begin{abstract}
Autonomous exploration in unknown environments requires unmanned aerial vehicles (UAVs) to efficiently generate informative trajectories while simultaneously constructing accurate maps. Although many existing exploration methods rely on geometric information, they often lack semantic awareness, resulting in suboptimal exploration efficiency and limited environmental understanding. 
To address this limitation, this paper proposes a semantic-aware exploration framework that adds semantic information to a roadmap-based exploration strategy. The proposed method builds on the Dynamic Exploration Planner (DEP), which incrementally constructs a Probabilistic Roadmap (PRM), and augments this roadmap with a semantic layer. 
A semantic reward function is introduced to prioritize regions containing meaningful objects and structures, enabling the UAV to focus on areas with higher information value. Furthermore, the roadmap is continuously updated to support efficient frontier selection and path planning during exploration. 
The proposed framework is implemented in ROS Noetic and Gazebo using an RGB-D sensor for simultaneous acquisition of geometric and semantic information. Experimental results in multiple simulated environments demonstrate that the proposed approach achieves exploration coverage rates between 90\% and 94\% while reducing exploration time and travel distance compared with conventional geometry-based exploration methods. 
\end{abstract}

\begin{IEEEkeywords}
UAV, autonomous exploration, semantic exploration, 3D mapping, probabilistic roadmap
\end{IEEEkeywords}

%%%%%%%%%%%%%%%%%%%%%%%%%%%%%%%%%%%%%%%%%%%%%%%%%%%%%%%%%%%%%%%%%%%%%%%%%%%%%%%%
\section{Introduction}
\label{sec:introduction}

Autonomous exploration is a core capability for unmanned aerial vehicles (UAVs) operating in unknown environments, with applications in search and rescue, infrastructure inspection, environmental monitoring, and warehouse management~\cite{kareem2026safe}. Efficient exploration requires the UAV to select informative viewpoints and generate safe trajectories that maximize map coverage while minimizing exploration cost~\cite{collignon2025search}. Early methods select the boundaries between known and unknown space as exploration targets \cite{shim2005autonomous, rasche2010combining}. Such frontier-based approaches are computationally efficient but degrade in large and complex three-dimensional environments. Sampling-based planners~\cite{selin2019efficient, zhou2021fuel, bircher2016receding, cao2021tare, huppi2022t, zhang2025fsmp, cao2024deep} instead evaluate candidate viewpoints by information gain. These planners generally provide better scalability and exploration performance in three-dimensional environments.

However, most of these methods rely solely on geometric information from occupancy or voxel maps~\cite{hornung2013octomap}, so every unexplored region is treated as equally important regardless of its relevance to the task. In practice, regions such as doorways, corridors, and object-rich areas often carry more useful information than ordinary free space, and ignoring this leads to redundant motion and inefficient exploration. Recent advances\cite{lu2024semantics, dharmadhikari2023semantics, papatheodorou2023finding, asgharivaskasi2023semantic} in semantic perception allow robots to extract such high-level context, and semantic-guided exploration has been shown to improve efficiency by directing the UAV toward semantically informative regions.

Motivated by these observations, this paper proposes a semantic-aware exploration framework for UAVs operating in unknown indoor environments. The proposed method extends the DEP by integrating semantic rewards into the viewpoint evaluation process. A voxel-based semantic map is constructed to represent both geometric and semantic information, while a PRM-based planner generates efficient trajectories between candidate viewpoints. By jointly considering geometric information gain and semantic importance, the proposed framework enables the UAV to explore unknown environments more effectively.
The main contributions of this work are summarized as follows:
\begin{itemize}
\item A semantic-aware exploration framework that integrates semantic information into a DEP-based exploration architecture.
\item A semantic reward mechanism that prioritizes exploration targets according to their contextual significance.
\item Evaluation in multiple indoor environments demonstrating improved exploration efficiency.
\end{itemize}

The remainder of this paper is organized as follows: Sec \ref{sec:related} presents the related works, while Sec \ref{sec:method} describes the proposed system for semantic-aware autonomous exploration. The experiments conducted and results analysis are presented in Sec \ref{sec:exp}. Finally, Sec \ref{sec:conclusion} draws a conclusion with future work.

\section{Related Work} \label{sec:related}

\subsection{Autonomous Exploration}
Autonomous exploration is broadly divided into frontier-based and sampling-based approaches. Frontier-based methods select the boundaries between known and unknown space as exploration targets \cite{yamauchi1997frontier}. They are efficient in two-dimensional settings but scale poorly to large three-dimensional environments. Sampling-based planners instead generate candidate viewpoints and rank them by expected information gain, which gives better flexibility and scalability in 3D. Representative approaches include the PRM~\cite{huppi2022t}, Rapidly-exploring Random Tree~\cite{luo2023survey}, Next-Best View (NBV)~\cite{bircher2016receding}, DEP~\cite{xu2021autonomous}, and Autonomous Exploration Planner (AEP)~\cite{selin2019efficient}. Later work adds historical reuse, trajectory optimization, and uncertainty-aware planning \cite{karaman2011sampling}, and recent UAV planners\cite{zhou2021fuel, cao2021tare, zhao2023autonomous, zhang2025fsmp, cao2024deep} improve speed and scalability through hierarchical and dual-stage planning, mixed frontier-sampling strategies, and learning-based viewpoint selection. However, these frameworks rely mainly on geometric information and treat all unexplored regions as equally important.

\subsection{Semantic-Aware Exploration}
Semantic perception lets robots reason about an environment beyond its geometry, and semantic mapping enriches occupancy maps with object labels and scene context \cite{de2021real}. Building on this, semantic-aware exploration prioritizes regions that contain task-relevant objects or landmarks \cite{dharmadhikari2023semantics, papatheodorou2023finding}, and recent object-centric and information-theoretic formulations search for objects of interest more efficiently \cite{lu2024semantics, asgharivaskasi2023semantic}. Most of these studies target ground robots or rely on computationally expensive perception \cite{de2021real}, and the use of semantic rewards within sampling-based UAV exploration remains limited \cite{xu2024heuristic}. Motivated by this gap, we integrate semantic rewards into a DEP-based architecture that combines semantic information with voxel-based mapping and PRM-based planning to improve exploration efficiency in unknown indoor environments \cite{huppi2022t, hornung2013octomap}.

%===============================================================================

%===============================================================================

\section{Proposed Method} \label{sec:method}

\begin{figure*}[t]
\centering
\begin{tikzpicture}[
  modbox/.style={rectangle, rounded corners=3pt, draw, very thick, align=center,
                 inner sep=3pt, text width=2.5cm, font=\scriptsize, minimum height=1.0cm},
  iobox/.style ={rectangle, rounded corners=3pt, draw=gray!60, fill=gray!8, very thick,
                 align=center, inner sep=3pt, text width=1.85cm,
                 font=\scriptsize\bfseries, text=gray!25!black, minimum height=0.72cm},
  mapmod/.style ={modbox, fill=Cenv!10,    draw=Cenv,      text=Cenv!88!black},
  semmod/.style ={modbox, fill=Ccritic!12, draw=Ccritic,   text=Ccritic!88!black},
  roadmod/.style={modbox, fill=Cactor!12,  draw=Cactor,    text=Cactor!92!black},
  loopmod/.style={modbox, fill=Cbuf!12,    draw=Cbuf,      text=Cbuf!88!black},
  trajmod/.style={modbox, fill=Cpd!10,     draw=Cpd!85,    text=Cpd!85!black},
  Aarr/.style ={-{Stealth[length=5pt,width=3.4pt]}, thick, gray!40!black},
  Abi/.style={{Stealth[length=5pt,width=3.4pt]}-{Stealth[length=5pt,width=3.4pt]}, thick, gray!40!black},
  lbl/.style ={font=\scriptsize, text=gray!45!black, inner sep=1.5pt}]
 
  \node[iobox] (pose)  at (0, 2.45) {Odometry /\\ Pose};
  \node[iobox] (cloud) at (0, 0.45) {Point Cloud};
 
  \node[mapmod] (map) at (2.9, 1.45)
    {\textbf{\small Map Module}\\[1pt]3D occupancy (log-odds)\\ESDF};
 
  \node[semmod]  (sem)  at (7.3, 2.45)
    {\textbf{\small Semantic Module}\\[1pt]detect \& filter\\create regions};
  \node[roadmod] (road) at (11.4, 2.45)
    {\textbf{\small Roadmap Mgmt.}\\[1pt]sample \& connect\\score \& update};
  \node[loopmod] (loop) at (9.35, 0.85)
    {\textbf{\small Main Loop}\\[1pt]check goal / safety\\trigger replanning};
 
  \begin{scope}[on background layer]
    \node[draw=Cpborder, very thick, dashed, rounded corners=5pt,
          fill=Cpanel, inner sep=10pt, fit=(sem)(road)(loop)] (planner) {};
  \end{scope}
  \node[font=\scriptsize\bfseries, text=Cpborder!55!black, anchor=north west]
        at ([xshift=2pt]planner.north west) {Exploration Planner};
 
  \node[trajmod] (traj) at (9.35, -1.25)
    {\textbf{\small Trajectory}\\[1pt]generate path\\ESDF optimization};
  \node[iobox] (ctrl) at (13.7, -1.25) {Control\\ Signal};
 
  \draw[Aarr] (pose)  -- (map);
  \draw[Aarr] (cloud) -- (map);
  \draw[Aarr] (pose.east) -- node[lbl, above]{pose} (sem.west);
  \draw[Aarr] (map.east) -- node[lbl, above]{3D map} (map.east -| planner.west);
  \draw[Aarr] (sem) -- node[lbl, above, align=center]{semantic\\ regions} (road);
  \draw[Abi] (road.south) |- (loop.east);
  \draw[Aarr] (loop) -- node[lbl, right]{reference path} (traj);
  \draw[Aarr] (map.south) |- node[lbl, below, pos=0.72]{ESDF} (traj.west);
  \draw[Aarr] (traj) -- node[lbl, above]{control} (ctrl);
\end{tikzpicture}
\caption{Overview of the proposed semantic-aware exploration framework. Sensor inputs build the 3D map, which feeds the exploration planner.}
\label{fig:system_overview}
\end{figure*}
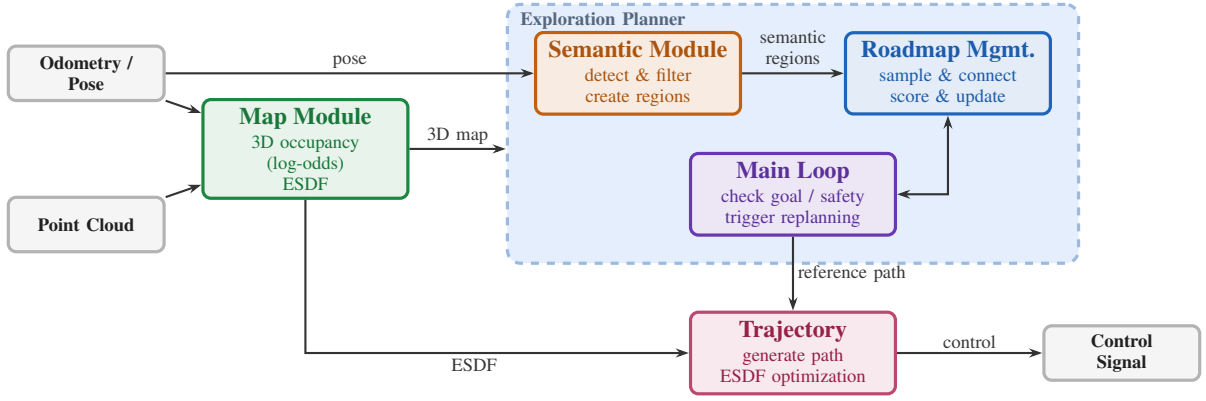

The proposed framework integrates semantic information into a roadmap-based exploration strategy to improve exploration efficiency in unknown environments. As illustrated in Fig.~\ref{fig:system_overview}, the system consists of three main modules: the Map Module, Semantic Module, and Roadmap Management Module. These modules operate in a closed-loop manner and continuously exchange information during exploration.

\subsection{Problem Formulation}

Consider an unknown indoor environment $V_b \subset \mathbb{R}^{3}$, where a UAV equipped with an RGB-D sensor is deployed for autonomous exploration. The environment representation follows standard volumetric mapping approaches \cite{hornung2013octomap}. The environment consists of free space $V_{free}$, occupied space $V_{occ}$, and unexplored space $V_{un}$ with $V_b = V_{free} \cup V_{occ} \cup V_{un}$. Initially, only a local region around the UAV, denoted by $V_{init}$, is known, while the remaining environment remains unexplored.

The objective of autonomous exploration is to maximize environment coverage while minimizing exploration cost. Given the current occupancy map $M$, the UAV incrementally samples candidate viewpoints and constructs a PRM for navigation. Following the DEP~\cite{bircher2016receding}, the information gain associated with a candidate viewpoint $n$ is computed as in:

\begin{equation}
G(n) = w_nU_n + w_fU_f + w_sU_s,
\label{eq:gain_geo}
\end{equation}
where $U_n$, $U_f$, and $U_s$ denote the numbers of normal voxels, frontier voxels, and surface voxels observable from viewpoint $n$, respectively, and $w_n$, $w_f$, and $w_s$ are their corresponding weighting factors. At each exploration iteration, the UAV selects the viewpoint with the highest utility and generates a collision-free trajectory through the roadmap. The exploration process is considered complete when all reachable regions have been mapped:

\begin{equation}
V_{mapped} = \left( V_{free} \cup V_{occ} \right) \setminus V_{unreachable}.
\end{equation}

%%%%%%%%%%%%%%%%%%%%%%%%%%%%%%%%%%%%%%%%%%%%%%%%%%%%%%%%%%

\subsection{Map Module}

The Map Module is responsible for constructing and maintaining a three-dimensional occupancy representation of the environment from incoming point cloud measurements. Inspired by~\cite{asgharivaskasi2023semantic,xu2021autonomous}, the occupancy state of each voxel is represented using the log-odds formulation:
\begin{equation}
\ell(v_i)=
\log \frac{p(v_i)}{1-p(v_i)},
\end{equation}

As new point cloud data arrive, ray-casting is performed to update free and occupied voxels within the volumetric map representation, as commonly adopted in octree-based mapping frameworks~\cite{hornung2013octomap}. Simultaneously, an Euclidean Signed Distance Field is generated from the occupancy map to provide obstacle distance information for safe trajectory generation.

%%%%%%%%%%%%%%%%%%%%%%%%%%%%%%%%%%%%%%%%%%%%%%%%%%%%%%%%%%

\subsection{Semantic Module}
Conventional exploration methods evaluate candidate viewpoints solely according to geometric information gain. However, not all unexplored regions contribute equally to the exploration objective. Regions containing entrances, corridors, intersections, or task-relevant objects often provide higher contextual value than ordinary free space. The Semantic Module introduces task-oriented exploration by assigning importance to specific regions of interest, extending semantic-aware exploration strategies proposed in \cite{canh2024s3m}. For each detected semantic object, the system generates a semantic region:

\begin{equation}
S_i=(p_i,l_i,w_i,r_i),
\end{equation}
where \(p_i\) denotes the centroid position, \(l_i\) is the semantic label, \(w_i\) is the semantic importance weight, and \(r_i\) is the influence radius. Given a roadmap node \(n\) located at position \(p_n\), the distance to semantic region \(S_i\) is $d_i=\|p_n-p_i\|$. The semantic reward contributed by region \(S_i\) is defined as

\begin{equation}
R_i(n)=
\begin{cases}
w_i\left(1-\dfrac{d_i}{r_i}\right), & d_i \le r_i,\\
0, & d_i > r_i.
\end{cases}
\end{equation}

The total semantic reward of node \(n\) is then computed as

\begin{equation}
R_{\mathrm{sem}}(n) =
\sum_{i \in \mathcal{S}} R_i(n),
\label{eq:rsem}
\end{equation}

where $\mathcal{S}$ is the set of semantic regions. The reward decreases linearly with distance and vanishes outside the influence radius, so the UAV is encouraged to prioritize semantic targets while maintaining exploration efficiency, following information-driven exploration methods~\cite{selin2019efficient}.  The overall utility combines the geometric information gain and the semantic reward:

\begin{equation}
U(p) = G(p) + R_{\mathrm{sem}}(p).
\label{eq:utility}
\end{equation}

The exploration problem is therefore formulated as:

\begin{equation}
p^{*} = \arg\max_{p \in P} U(p),
\end{equation}

where $P$ denotes the set of candidate viewpoints. In practice, this utility is divided by the estimated travel cost during viewpoint selection, so that nearby high-value viewpoints are preferred. By jointly considering geometric information gain and semantic importance, the proposed framework enables the UAV to prioritize informative and semantically meaningful regions during exploration.

%%%%%%%%%%%%%%%%%%%%%%%%%%%%%%%%%%%%%%%%%%%%%%%%%%%%%%%%%%

\subsection{Roadmap Management Module}
\label{subsec:roadmap}

The roadmap management module is responsible for incrementally constructing and maintaining a navigation graph that guides the UAV toward informative exploration targets. Inspired by incremental sampling-based exploration strategies \cite{xu2021autonomous, xu2024heuristic} and classical roadmap construction methods \cite{karaman2011sampling}, the roadmap is continuously updated as new observations become available. At each planning cycle, the current UAV position is inserted into the existing roadmap and serves as the root node for subsequent graph expansion. A set of candidate nodes is then sampled within the unexplored and traversable space, following frontier-based exploration principles. To guarantee flight safety, each sampled node must satisfy collision-free constraints and lie inside the sensor-observable region, which is typically derived from volumetric mapping frameworks. These nodes are subsequently connected to nearby roadmap vertices if the connecting edge is free of obstacles.

\begin{figure*}[t]
\centering
% coordinates of the roadmap nodes (local to this figure)
\def\cA{(0.55,0.55)} \def\cB{(1.35,0.78)} \def\cC{(2.15,0.6)}
\def\cD{(2.85,1.05)} \def\cE{(2.45,1.70)} \def\cF{(1.65,1.55)}
\def\cG{(0.95,1.25)} \def\cH{(1.95,2.45)} \def\cI{(2.75,2.30)}
\def\cJ{(1.40,2.10)}
\def\sA{(2.25,2.05)} \def\sB{(2.55,2.45)} \def\sC{(2.95,1.65)} \def\sD{(1.15,2.15)}
\def\unkregion{%
  \fill[black!20] (0,3.4)--(1.7,3.4)--(1.7,2.4)--(0.75,2.4)--(0.75,1.45)--(0,1.45)--cycle;
  \draw[black!45,thick] (0,3.4)--(1.7,3.4)--(1.7,2.4)--(0.75,2.4)--(0.75,1.45)--(0,1.45);}
\def\basegraph{%
  \unkregion
  \draw[Cenv!65,thick] \cA--\cB \cB--\cC \cC--\cD \cD--\cE \cE--\cF \cF--\cG
       \cG--\cA \cF--\cH \cH--\cI \cI--\cE \cH--\cJ \cJ--\cF;
  \foreach \p in {\cB,\cC,\cD,\cE,\cF,\cG,\cH,\cI}
     \filldraw[fill=Cenv!85,draw=Cenv!55!black,line width=0.4pt] \p circle(1.7pt);
  \draw[red!75!black, line width=1.5pt] \cA--\cG \cG--\cF \cF--\cJ;
  \filldraw[fill=red!75!black,draw=red!50!black] \cA circle(2.3pt);
  \draw[Cpd!75!black, line width=0.5pt] \cJ circle(4.3pt);
  \filldraw[fill=Cpd,draw=Cpd!55!black,line width=0.4pt] \cJ circle(2.5pt);}
% updated roadmap (result of panel (c)) drawn green; reused in panel (d)
\def\updatededges{%
  \draw[Cenv!65,thick] \sA--\cH \sA--\cE \sB--\cI \sB--\sA \sC--\cE \sC--\cD \sD--\cJ \sD--\cG;
  \foreach \p in {\sA,\sB,\sC,\sD}
     \filldraw[fill=Cenv!85,draw=Cenv!55!black,line width=0.4pt] \p circle(1.7pt);}
\resizebox{\textwidth}{!}{%
\begin{tikzpicture}[
  panel/.style={draw=gray!45, thin, rounded corners=2pt},
  plbl/.style={font=\scriptsize, text=gray!25!black, inner sep=1.2pt},
  subcap/.style={font=\small},
  semstar/.style={star, star points=5, star point ratio=2.3, fill=Csem,
                  draw=Csem!55!black, line width=0.4pt, inner sep=0pt, minimum size=11pt},
  rewring/.style={draw=Csem, line width=1.1pt},
  semcirc/.style={draw=Csem, dashed, line width=0.8pt},
  guide/.style={draw=Csem, line width=2.2pt, opacity=0.4, line cap=round,
                -{Stealth[length=3.6pt,width=3pt]}}]
  \begin{scope}[shift={(0,0)}]
    \draw[panel] (0,0) rectangle (3.4,3.4);
    \basegraph
    \node[plbl, fill=black!20, inner sep=1pt] at (0.8,3.02) {unknown};
    \node[plbl] at (0.55,0.2) {start};
  \end{scope}
  \node[subcap] at (1.7,-0.42) {(a)};
  \begin{scope}[shift={(4.45,0)}]
    \draw[panel] (0,0) rectangle (3.4,3.4);
    \basegraph
    \foreach \p in {(2.95,2.85),(2.55,2.45),(2.95,1.65),(2.25,2.05),%
                    (1.15,2.15),(1.70,2.32),(3.02,2.18),(2.10,2.85)}
       \fill[Ccritic] \p circle(1.6pt);
    \node[plbl, text=Ccritic!85!black, anchor=west] at (1.78,2.95) {new samples};
  \end{scope}
  \node[subcap] at (6.15,-0.42) {(b)};
  \begin{scope}[shift={(8.9,0)}]
    \draw[panel] (0,0) rectangle (3.4,3.4);
    \basegraph
    \foreach \p in {(2.95,2.85),(1.70,2.32),(2.10,2.85),(3.02,2.18)}
       \fill[Ccritic!35] \p circle(1.4pt);
    \draw[Cactor,thick] \sA--\cH \sA--\cE \sB--\cI \sB--\sA \sC--\cE \sC--\cD \sD--\cJ \sD--\cG;
    \foreach \p in {\sA,\sB,\sC,\sD}
       \filldraw[fill=cyan!75,draw=Cactor!70!black,line width=0.4pt] \p circle(1.8pt);
    \node[plbl, text=Cactor!85!black, anchor=west] at (1.95,2.95) {new edges};
  \end{scope}
  \node[subcap] at (10.6,-0.42) {(c)};
  % ---- (d) semantic layer ----
  \begin{scope}[shift={(13.35,0)}]
    \draw[panel] (0,0) rectangle (3.4,3.4);
    \basegraph
    \updatededges
    \def\sem{(2.6,1.45)}
    \draw[guide] \cJ .. controls (2.0,1.95) .. \sem;
    \draw[semcirc] \sem circle(0.62);
    \foreach \p in {\cD,\cE,\sC} \draw[rewring] \p circle(0.13);
    \node[semstar] at \sem {};
    \node[plbl, text=Csem!85!black, anchor=west] at (1.75,2.98) {semantic reward};
  \end{scope}
  \node[subcap] at (15.05,-0.42) {(d)};
  % ---- legend for the new semantic glyphs ----
  \begin{scope}[shift={(0,-0.8)}]
    \node[semstar, scale=0.85] at (0.4,0) {};
    \node[anchor=west, font=\scriptsize] at (0.62,0) {semantic node $S_i$};
    \draw[semcirc] (4.0,0) circle(0.13);
    \node[anchor=west, font=\scriptsize] at (4.25,0) {influence radius $r_i$};
    \filldraw[fill=Cenv!85,draw=Cenv!55!black,line width=0.4pt] (7.6,0) circle(1.7pt);
    \draw[rewring] (7.6,0) circle(0.13);
    \node[anchor=west, font=\scriptsize] at (7.85,0) {semantically rewarded node};
    \draw[Csem, line width=2pt, -{Stealth[length=3.6pt,width=3pt]}] (12.4,0)--(12.95,0);
    \node[anchor=west, font=\scriptsize] at (13.05,0) {guided exploration};
  \end{scope}
\end{tikzpicture}}
\caption{Incremental roadmap update process. Green nodes and edges form the current roadmap, the thick red path is the previously executed trajectory, the magenta marker is the current UAV pose, and the gray region is unexplored. (a)~The roadmap after the previous trajectory is completed. (b)~New candidate nodes (orange) are sampled in the unexplored, traversable space. (c)~Valid candidates (cyan) are connected to the roadmap by new collision-free edges (blue), giving the updated graph for the next planning cycle. (d)~The semantic layer is then applied to the updated roadmap. Each detected object defines a semantic node $S_i$ (purple star) with an influence radius $r_i$ (dashed circle). Roadmap nodes inside this radius receive a semantic reward (purple ring), which biases viewpoint selection toward semantically relevant regions (arrow). The UAV is therefore guided toward objects of interest in addition to geometric frontiers.}
\label{fig:roadmap_update}
\end{figure*}
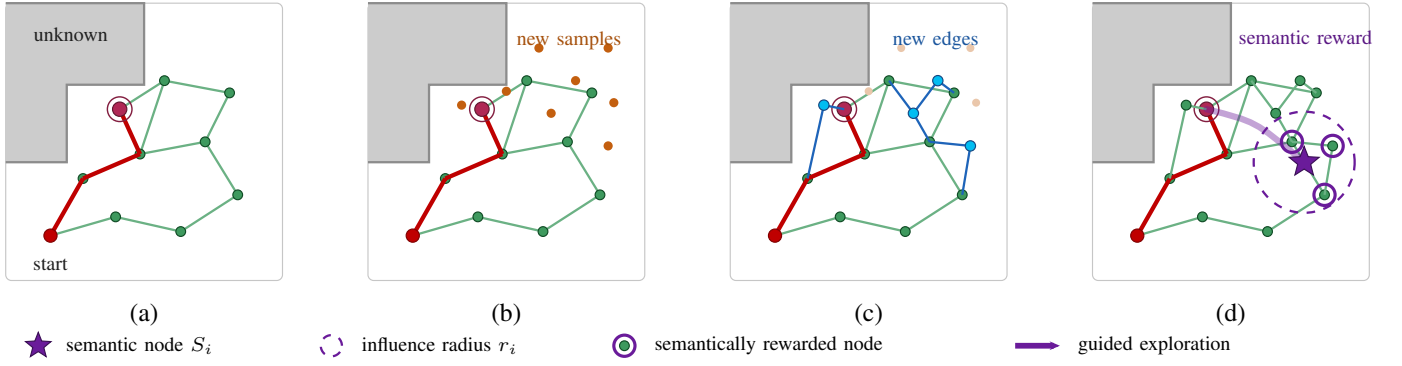

Figure~\ref{fig:roadmap_update} illustrates the incremental roadmap update process. Figure~\ref{fig:roadmap_update}(a) shows the roadmap after completing the previous trajectory. Figure~\ref{fig:roadmap_update}(b) presents newly sampled nodes generated in the unexplored region. Finally, Figure~\ref{fig:roadmap_update}(c) depicts the insertion of valid nodes and the creation of new collision-free edges, resulting in an updated roadmap for future planning. After the roadmap has been updated, each candidate node is evaluated according to its expected exploration utility. The geometric information gain $G(n_i)$ of node $n_i$ is computed from the numbers of observable normal, frontier, and surface voxels, as defined in Eq.~\eqref{eq:gain_geo}.

The semantic reward $R_{\mathrm{sem}}(n_i)$ of the same node is obtained from Eq.~\eqref{eq:rsem}, which sums the contributions of all semantic regions whose influence radius covers the node. The travel cost associated with reaching node $n_i$ is estimated as

\begin{equation}
T_{\mathrm{path}}
=
\frac{d_i}{v}
+
w_{\psi}
\frac{\Delta \psi_i}{\omega},
\label{eq:path_cost}
\end{equation}
where $d_i$ is the path length, $v$ is the UAV translational velocity, $\Delta \psi_i$ is the required yaw rotation, $\omega$ is the angular velocity, and $w_{\psi}$ is a weighting coefficient. The final utility score of node $n_i$ is the per-node form of the utility $U$ in Eq.~\eqref{eq:utility}, normalized by the travel cost:

\begin{equation}
Score_i
=
\frac{
G(n_i)
+
R_{\mathrm{sem}}(n_i)
}
{
T_{\mathrm{path}}
}.
\label{eq:node_score}
\end{equation}

Nodes with higher exploration gain, stronger semantic relevance, and lower traversal cost obtain larger scores and are therefore preferred as exploration targets. The node evaluation procedure is summarized in Algorithm~\ref{alg:semantic_node_evaluation}.

\begin{algorithm}[t]
\caption{Semantic Node Evaluation}
\label{alg:semantic_node_evaluation}
\KwIn{Roadmap $\mathcal{R}$, semantic regions $\{S_k\}$}
\KwOut{Node scores $\{Score_i\}$}

\For{node $n_i \in \mathcal{R}$}
{
    $G(n_i) \leftarrow w_n U_n + w_f U_f + w_s U_s$\;

    $R_{\mathrm{sem}}(n_i) \leftarrow 0$\;

    \For{semantic region $S_k$}
    {
        $d_k \leftarrow \|p_i-p_k\|$\;

        \If{$d_k \le r_k$}
        {
            $R_{\mathrm{sem}}(n_i)
            \leftarrow
            R_{\mathrm{sem}}(n_i)
            +
            w_k
            \left(
            1-\frac{d_k}{r_k}
            \right)$\;
        }
    }

    $T_{\mathrm{path}}
    \leftarrow
    \dfrac{d_i}{v}
    +
    w_{\psi}
    \dfrac{\Delta\psi_i}{\omega}$\;

    $Score_i
    \leftarrow
    \dfrac{
    G(n_i) + R_{\mathrm{sem}}(n_i)}
    {T_{\mathrm{path}}}$\;
}

\Return{$\{Score_i\}$}
\end{algorithm}

\section{Experimental Setup} \label{sec:exp}
 
\subsection{Simulation Environments}

\begin{figure}
    \centering
    \includegraphics[width=0.7\linewidth]{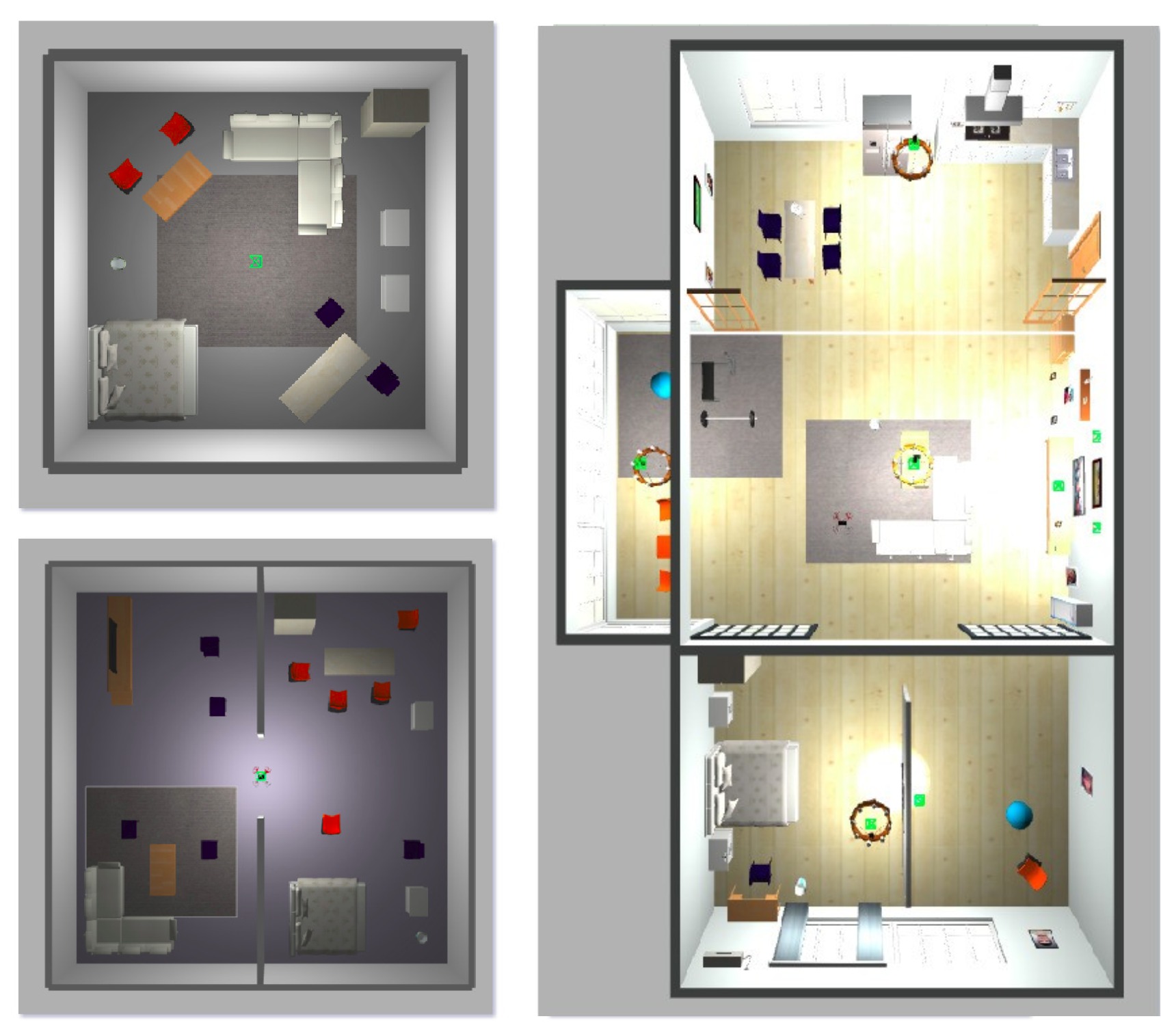}
    \caption{The three simulated indoor environments used for evaluation, ordered from left to right by increasing size and complexity.}

    \label{fig:env}
\end{figure}

To evaluate the performance of the proposed method under different levels of environmental complexity, three indoor environments were constructed in Gazebo with ROS Noetic. The environments differ in size, room layout, obstacle density, and visibility constraints, as shown in Fig.~\ref{fig:env}. Environment~1 represents a small single-room apartment with a size of 50m$^2$, Environment~2 contains two interconnected rooms separated by a narrow passage with a size of 100m$^2$, and Environment~3 represents a large apartment with multiple connected rooms and corridors with a size of 200m$^2$. 
A quadrotor UAV was employed in all experiments. The UAV was equipped with an RGB-D camera and an IMU sensor for perception and state estimation. The sensor specifications used throughout the experiments are listed in Table~\ref{tab:uav_sensor}.

\subsection{Sampling Density Analysis}

The performance was assessed using four metrics: total flight time (Time\_f, $\downarrow$), traveled distance (Distance, $\downarrow$), map coverage (Coverage, $\uparrow$), and the time required to reach $85\%$ coverage (Time\_r, $\downarrow$). Table~\ref{tab:sampling_comparison} presents the influence of sampling density on exploration performance. In Environment~1, the roadmap with 30 sampled nodes achieved the best trade-off between exploration efficiency and map coverage, completing exploration in 81.22~s over a path of 20.81~m while covering 91.23\% of the environment. Although the 25-node configuration reached a slightly higher final coverage of 91.89\%, it required a substantially longer exploration time of 130.41~s and a longer travel distance of 30.83~m. Increasing the number of samples beyond 35 nodes did not improve coverage and instead introduced additional computational overhead.

\begin{table}[!h]
\centering
\caption{UAV and sensor parameters used in simulation.}
\label{tab:uav_sensor}
\begin{tabularx}{\linewidth}{
   >{\centering\arraybackslash}X||
   >{\centering\arraybackslash}X}
\toprule
\textbf{Parameter} & \textbf{Value} \\ 
\midrule
UAV size & $0.5 \times 0.5 \times 0.3$ m \\ 
Camera type & RGB-D camera \\ 
Image resolution & $640 \times 480$ pixels \\ 
Frame rate & 30 FPS \\
Horizontal FOV & $90^\circ$ \\ 
Vertical FOV & $90^\circ$ \\ 
Depth range & $0.5 - 5.0$ m \\ 
Focal length & $f_x = f_y = 554.25$ \\ 
IMU noise density & $1.7\times10^{-2}\ \mathrm{m/s^2/\sqrt{Hz}}$ \\ 
Gyroscope noise density & $1.7\times10^{-2}\ \mathrm{rad/s/\sqrt{Hz}}$ \\ 
\bottomrule
\end{tabularx}
\end{table}

\begin{table}[t]
\centering
\scriptsize
\caption{Evaluation of random sampling density in three experimental environments. ``N/A'' indicates that the $85\%$ coverage threshold was not reached. The \best{best} and \second{second-best} values of each metric among the settings.}
\label{tab:sampling_comparison}
\begin{tabularx}{\linewidth}{
   >{\centering\arraybackslash}m{0.016\textwidth} 
   >{\centering\arraybackslash}m{0.035\textwidth}
   >{\centering\arraybackslash}X
   >{\centering\arraybackslash}m{0.08\textwidth}
   >{\centering\arraybackslash}m{0.09\textwidth}
   >{\centering\arraybackslash}X}
\toprule
\textbf{Env.} & \textbf{no. S.} & \textbf{Time\_f (s)} & \textbf{Distance (m)} & \textbf{Coverage (\%)} & \textbf{Time\_r (s)} \\
\midrule
\multirow{5}{*}{1}
& 25 & 130.41 & 30.83 & \best{91.89} & 102.09 \\
& 30 & \best{81.22}  & \best{20.81} & \second{91.23} & \second{79.01} \\
& 35 & \second{89.12}  & \second{21.43} & 90.78 & \best{72.01} \\
& 40 & 107.99 & 24.53 & 90.94 & 93.94 \\
& 45 & 62.18  & 8.09  & 72.36 & N/A \\
\midrule
\multirow{5}{*}{2}
& 25 & 225.24 & 57.08 & \second{95.75} & 146.01 \\
& 30 & \best{128.72} & \second{36.56} & 92.76 & \second{107.02} \\
& 35 & \second{129.04} & \best{35.25} & 93.78 & \best{95.03} \\
& 40 & 225.78 & 47.19 & \best{96.01} & 148.01 \\
& 45 & 228.03 & 48.55 & 93.86 & 153.04 \\
\midrule
\multirow{5}{*}{3}
& 30 & 262.11 & 70.01 & 79.89 & N/A \\
& 35 & 329.52 & \best{77.63} & 87.66 & 217.27 \\
& 40 & \second{310.04} & \second{78.04} & 89.06 & \second{190.34} \\
& 45 & \best{293.12} & 79.93 & \best{90.42} & \best{176.42} \\
& 50 & 351.69 & 83.93 & \second{89.85} & 224.03 \\
\bottomrule
\end{tabularx}
\end{table}

A similar tendency can be observed in Environment~2, where the 35-node configuration reached the $85\%$ coverage threshold in the shortest time of 95.03~s while maintaining a high final coverage of 93.78\% and the shortest travel distance of 35.25~m. For Environment~3, which contains more complex spatial structures, a denser roadmap was required. The 45-node configuration achieved the highest overall exploration efficiency, reaching the threshold in 176.42~s and attaining the highest coverage of 90.42\% among all settings, and was therefore selected for the subsequent comparison experiments.

These results indicate that an appropriate sampling density is essential for maintaining roadmap connectivity while avoiding redundant computations. Excessive sampling increases planning complexity, whereas insufficient sampling degrades exploration performance.

\subsection{Quantitative and Qualitative Results}
\begin{figure*}[!ht]
    \centering
    \begin{overpic}[width=0.9\textwidth]{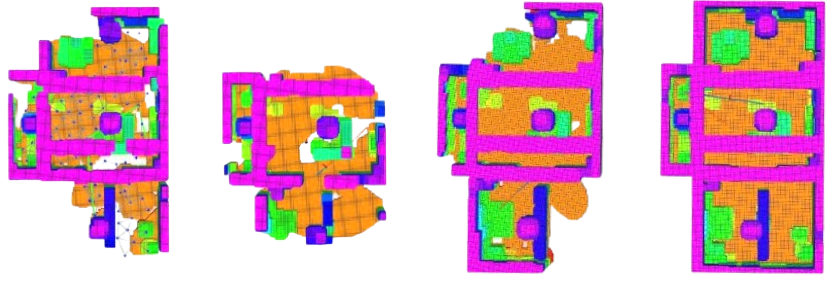}
         \put(8,0){AEP\cite{selin2019efficient}}
         \put(34,0){NBV\cite{bircher2016receding}}
         \put(60,0){DEP\cite{xu2021autonomous}}
         \put(83,0){\textbf{Our Proposed}}
    \end{overpic}
    \caption{Qualitative results comparison in Environment 3.}
    \label{fig:qual_env3}
\end{figure*}
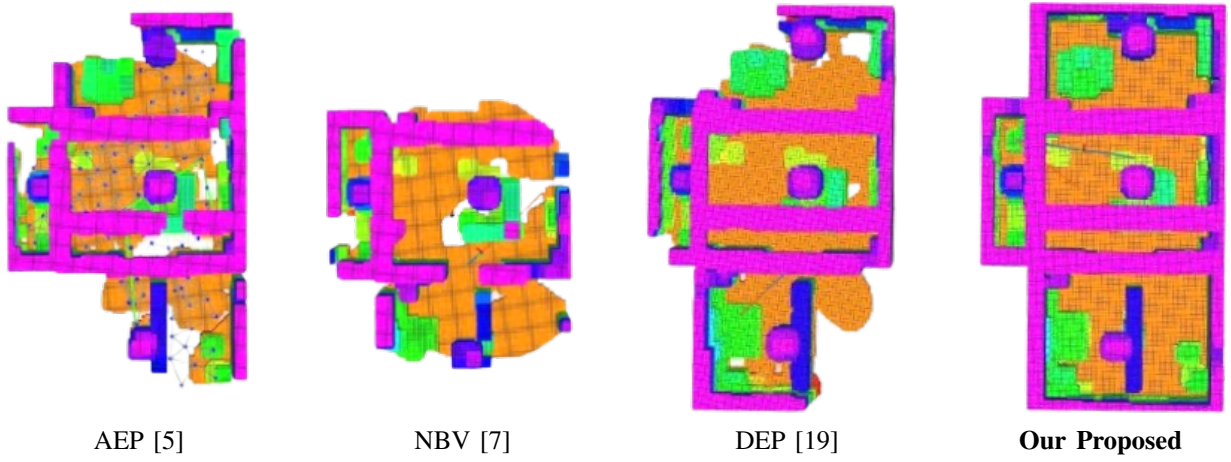
\begin{table}[!h]
\centering
\caption{Exploration performance across different environments. For each environment, \best{best} and \second{second-best} values are highlighted among the methods that complete exploration (DEP, AEP, and the proposed method). NBV is reported for reference but not ranked, since it attains the lowest coverage in every environment and its shorter time and distance result from incomplete exploration.}
\label{tab:DGMT_all_style_reduced}
\scriptsize
\begin{tabularx}{\linewidth}{
   >{\centering\arraybackslash}m{0.02\textwidth} ||
   >{\centering\arraybackslash}X|
   >{\centering\arraybackslash}X|
   >{\centering\arraybackslash}m{0.08\textwidth}|
   >{\centering\arraybackslash}m{0.09\textwidth}}
\toprule
\textbf{Env.} & \textbf{Algorithm} &
\textbf{Time\_f (s)} & \textbf{Distance (m)} & \textbf{Coverage (\%)} \\
\midrule
\multirow{4}{*}{1}
& NBV~\cite{bircher2016receding} & 115.69 & 11.84 & 82.70 \\
& AEP~\cite{selin2019efficient} & 134.71 & 24.69 & 83.10 \\
& DEP~\cite{xu2021autonomous} & \second{91.53} & \second{21.47} & \second{83.53} \\
& \textbf{Our Proposed} & \best{81.22} & \best{20.81} & \best{91.23} \\
\midrule
\multirow{4}{*}{2}
& NBV~\cite{bircher2016receding} & 150.71 & 15.56 & 81.06 \\
& AEP~\cite{selin2019efficient} & 194.68 & \second{45.72} & 92.04 \\
& DEP~\cite{xu2021autonomous} & \second{182.72} & 51.99 & \second{94.03} \\
& \textbf{Our Proposed} & \best{129.04} & \best{35.25} & \best{94.18} \\
\midrule
\multirow{4}{*}{3}
& NBV~\cite{bircher2016receding} & 209.35 & 20.96 & 51.11 \\
& AEP~\cite{selin2019efficient} & 443.23 & 123.46 & 79.75 \\
& DEP~\cite{xu2021autonomous} & \second{351.69} & \second{95.37} & \second{83.27} \\
& \textbf{Our Proposed} & \best{293.12} & \best{79.93} & \best{90.42} \\
\bottomrule
\end{tabularx}
\end{table}

The quantitative comparison is summarized in Table~\ref{tab:DGMT_all_style_reduced}. In Environment~1, the proposed framework achieved the shortest exploration duration of 81.22~s while obtaining the highest map coverage of 91.23\%. Compared with DEP, NBV, and AEP, the exploration time was reduced by approximately 11.3\%, 29.8\%, and 39.7\%, respectively. In Environment~2, the proposed method maintained a comparable coverage level of 94.18\% while significantly reducing the exploration duration to 129.04~s. Compared with DEP and AEP, the exploration time was reduced by 29.4\% and 33.7\%, respectively. Although NBV generated shorter paths, its coverage was considerably lower (81.06\% against 94.18\%), indicating inefficient exploration behavior.
The largest improvement was observed in Environment~3. The proposed framework achieved a coverage of 90.42\%, outperforming DEP, NBV, and AEP by 7.15, 39.31, and 10.67 percentage points, respectively. Furthermore, the exploration duration was reduced by 16.7\% compared with DEP and by 33.9\% compared with AEP. Figure~\ref{fig:qual_env3} shows the qualitative comparison in this environment, where the proposed method explores the connected rooms and corridors more completely than the baselines.

\begin{figure}[!h]
    \centering
    \includegraphics[width=0.8\linewidth]{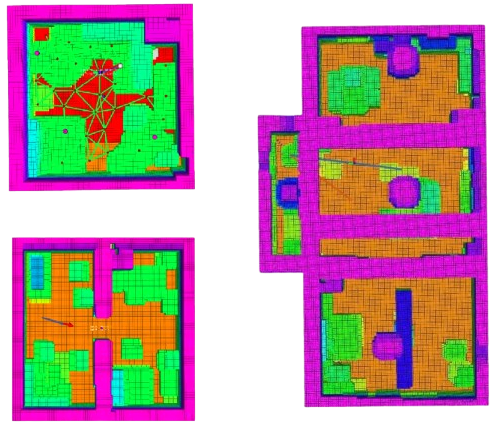}
    \caption{Qualitative results of the proposed method in the three environments.}
    \label{fig:qual_all}
\end{figure}

The superior performance of the proposed framework can be attributed to the integration of semantic information into the viewpoint evaluation process. By assigning higher priorities to semantically meaningful regions, the planner is able to discover informative areas earlier and reduce unnecessary exploration. In addition, the incremental roadmap management strategy preserves previously generated roadmap structures, reducing redundant sampling and improving planning efficiency.
Overall, the experimental results, together with the qualitative exploration maps in Fig.~\ref{fig:qual_all}, demonstrate that the proposed method consistently achieves higher map coverage and shorter exploration times than existing exploration approaches across different environments.

%===============================================================================

\section{Conclusion}
\label{sec:conclusion}

This paper presented a semantic-aware autonomous exploration framework for UAVs operating in unknown indoor environments. The proposed approach integrates semantic information into the viewpoint evaluation process, enabling the UAV to prioritize exploration targets that are not only geometrically informative but also semantically meaningful. In addition, an incremental roadmap management strategy was employed to maintain roadmap connectivity and improve planning efficiency throughout the exploration process. Experimental results demonstrated that the integration of semantic rewards significantly improved exploration performance compared with existing exploration methods. Future work will focus on integrating real-time semantic perception using deep learning models and validating the proposed framework on real UAV platforms. Furthermore, adaptive semantic weighting strategies and multi-UAV cooperative exploration will be investigated to further improve exploration efficiency in large-scale and dynamic environments.
%===============================================================================

% \section*{Acknowledgment}
% This work was supported by JST SPRING, Japan Grant Number JPMJSP2102.

\bibliographystyle{IEEEtran}
\bibliography{ref}  % .bib

\end{document}